\def\year{2020}\relax
\documentclass[letterpaper]{article} 
\usepackage{aaai20}  
\usepackage{times}  
\usepackage{helvet} 
\usepackage{courier}  
\usepackage{url}  
\usepackage{graphicx} 
\usepackage{xspace} 
\usepackage{booktabs}
\usepackage{amsmath,amssymb,amsfonts,amsthm}
\usepackage{cleveref}
\usepackage{nicefrac}
\usepackage{dsfont}
\usepackage{microtype}
\usepackage{mathtools}
\usepackage{xfrac}
\usepackage{mathabx}
\usepackage{verbatim}
\usepackage[dvipsnames]{xcolor}
\usepackage{tikz}
\usepackage{algorithm}
\usepackage[noend]{algpseudocode}
\usepackage{subcaption}
\usepackage{enumitem}
\usepackage{accents}
\usepackage[export]{adjustbox}
\usepackage[binary-units]{siunitx}

\usepackage{flushend}

\newcommand{\real}[0]{\mathbb{R}}

\newcommand{\graph}[0]{\mathcal{G}}
\newcommand{\vertexSet}[0]{\mathcal{V}}
\newcommand{\edgeSet}[0]{\mathcal{E}}
\newcommand{\vertex}[0]{v}
\newcommand{\vertexGroup}[0]{V}

\newcommand{\vertexStart}[0]{\vertex_s}
\newcommand{\vertexGoal}[0]{\vertex_g}
\newcommand{\Path}[0]{\xi}

\newcommand{\fullWorldSpace}[0]{\mathcal{X}}

\newcommand{\sensorSet}[0]{\mathcal{S}}
\newcommand{\sensor}[0]{\sigma}
\newcommand{\worldState}[1]{\fullWorldSpace_{#1}}
\newcommand{\world}[0]{\fullWorldSpace_{\vertexSet}}
\newcommand{\worldSet}[0]{\fullWorldSpace^{\lvert \vertexSet \rvert}}
\newcommand{\meas}[0]{y}
\newcommand{\measSet}[0]{\mathcal{Y}}
\newcommand{\measFnDef}[0]{H}

\newcommand{\utilityFnDef}[0]{F}
\newcommand{\utilityFn}[2]{\utilityFnDef\left(#1, #2\right)}
\newcommand{\marginalGain}[2]{\Delta_\utilityFnDef\left(#1, #2\right)}

\title{Adaptive Informative Path Planning with Multimodal Sensing}

\author{Shushman Choudhury\thanks{Equal contribution.},\textsuperscript{\rm 1}
Nate Gruver\footnotemark[1],\textsuperscript{\rm 1}
Mykel J. Kochenderfer\textsuperscript{\rm 2}\\
\textsuperscript{\rm 1} Computer Science, Stanford University\\
\textsuperscript{\rm 2} Aeronautics \& Astronautics, Stanford University\\
shushman@stanford.edu,
ngruver@stanford.edu, 
mykel@stanford.edu}

\begin{document}

\maketitle

\begin{abstract}
Adaptive Informative Path Planning (AIPP) problems model an agent tasked with obtaining information subject to resource constraints in unknown, partially observable environments. 
Existing work on AIPP has focused on representing observations about the world as a result of agent movement. We formulate the more general setting where the agent may choose between different sensors at the cost of some energy, in addition to traversing the environment to gather information. We call this problem AIPPMS (MS for Multimodal Sensing). AIPPMS requires reasoning jointly about the effects of sensing and movement in terms of both energy expended and information gained. We frame AIPPMS as a Partially Observable Markov Decision Process (POMDP) and solve it with online planning. Our approach is based on the Partially Observable Monte Carlo Planning framework with modifications to ensure constraint feasibility and a heuristic rollout policy tailored for AIPPMS. We evaluate our method on two domains: a simulated search-and-rescue scenario and a challenging extension to the classic RockSample problem. We find that our approach outperforms a classic AIPP algorithm that is modified for AIPPMS, as well as online planning using a random rollout policy. 
\end{abstract}

\section{Introduction}
\label{sec:intro}

For various robotic applications such as search-and-rescue, terrain exploration, and environmental monitoring, an autonomous agent must gather information in uncertain environments with partial observability. The agent is equipped with multiple sensing modalities with which it receives noisy observations of the world. It also has limited energy, which is expended both by using the sensors and by movement. We call this the problem of Adaptive Informative Path Planning with Multimodal Sensing (AIPPMS). In such settings, there is a trade-off between exploring the environment and exploiting the current belief about the environment to visit informative locations. Our objective is to obtain an adaptive strategy that guides the agent from the start to the goal location while balancing this exploration-exploitation tradeoff and respecting
the energy budget constraint.

AIPPMS can model a variety of real-world problems. For instance, in a search-and-rescue mission, one or more robots must traverse unstructured terrain, detect the presence of survivors with noisy sensors and then decide which locations to visit in order to save as many survivors as possible within their onboard battery life~\cite{singh2009nonmyopic}. Other domains include monitoring sensitive ecosystems~\cite{das2015data} and deploying remote devices to cover an unknown landscape~\cite{binney2012branch}.

\begin{figure}
    \centering
    \includegraphics[width=0.95\columnwidth]{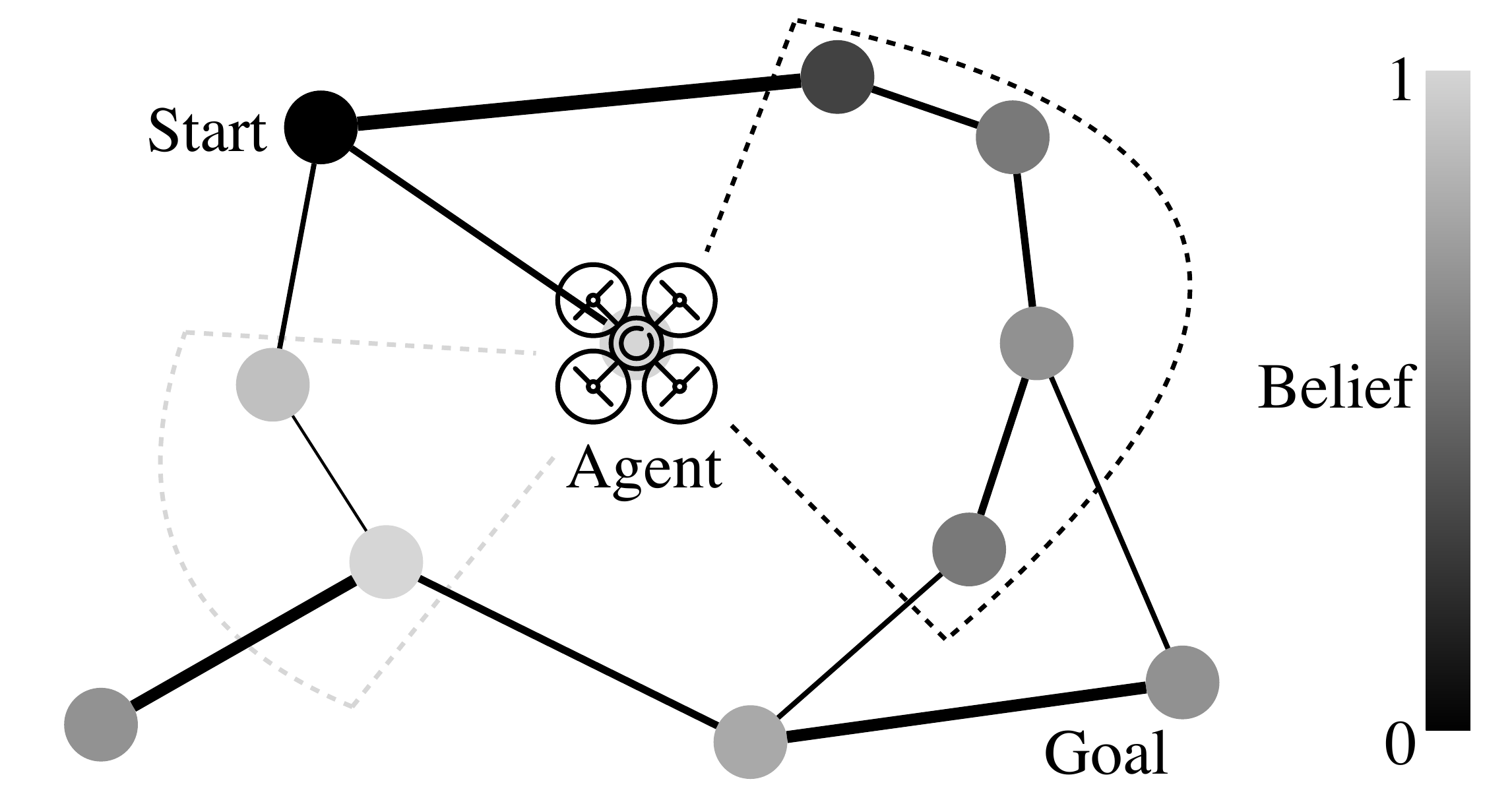}
    \caption{The AIPPMS problem requires an agent to plan paths over a graph of locations in an environment, while maximizing the information gathered by visiting locations subject to an energy constraint (the edge thickness represents the energy cost of traversal). The states of locations (binary in this example) are observed through noisy sensor readings. The agent is equipped with two kinds of sensors with different fidelity and range parameters. The shade of the node indicates the current belief of the agent about its state.}
    \label{fig:fig1}
\end{figure}

The informative path planning (IPP) problem is NP hard in the non-adaptive setting \cite{meliou2007nonmyopic}, where the agent plans and executes a path without accounting for noisy observations.  The adaptive setting adds another dimension of difficulty by requiring not just a plan but a policy that reacts to the observations the agent receives. We generalize the problem even further by explicitly introducing multiple sensing modalities and considering the effect of both sensing and moving on the energy budget.

Prior work has focused on non-adaptive IPP with various approximation algorithms~\cite{singh2007efficient}. The adaptive setting has been studied with utility function assumptions such as submodularity and locality~\cite{singh2009nonmyopic}, hypothesis identification in metric spaces~\cite{wee2016adaptive}, and a data distribution over possible environments~\cite{choudhury2017learning}. To the best of our knowledge, our multimodal sensing formulation has not been considered in previous work on adaptive IPP. Unlike our problem, in previous formulations the only decision is the location to visit next, observations are received on visiting locations, and only movement depletes energy.

Our key idea is to \emph{reason jointly about multimodal sensing and movement} by formulating the problem as a Partially Observable Markov Decision Process (POMDP) and using an online solver to accommodate the state and observation spaces, which are exponentially large (in the number of locations). We use constrained search in the online solver to ensure that the policy always satisfies the goal and energy constraints. For a tailored rollout policy to guide the online simulations, we use an adaptive extension to a near-optimal algorithm for the fully observable variant of our problem~\cite{zhang2016submodular}. Through this online POMDP framework, we obtain an adaptive IPP algorithm that judiciously balances exploration with multimodal sensing and exploitation with movement.
Our approach consistently outperforms a nonmyopic adaptive IPP algorithm~\cite{singh2009nonmyopic} and our ablation study shows that the tailored rollout policy is key to this improvement.

\section{Background}
\label{sec:related}
This section provides relevant background on the Informative Path Planning problem, Partially Observable Markov Decision Processes (POMDPs), and online POMDP solvers.

\subsection{Informative Path Planning}
\label{sec:related-ipp}

In informative path planning (IPP) we must choose the best subset of locations to visit (thereby gathering information) subject to constraints on the path, such as energy expended. The IPP problem is relevant to various mobile robotics applications such as mapping with terrestrial and aerial vehicles~\cite{heng2015efficient,charrow2015information} and adaptive sampling for underwater environments~\cite{binney2010informative}.

The IPP problem is NP-hard, as is path planning in general~\cite{canny1987complexity}. IPP can be framed as an orienteering problem~\cite{golden1987orienteering}, which is a generalization of the known NP-hard Traveling Salesman Problem with an additional constraint on how far the agent can travel. A number of efficient heuristics and approximation techniques have been explored. Gaussian Processes were used to model the environment and the mutual information between locations~\cite{singh2009efficient}, and minimum-cost tours have been computed on efficiently chosen subsets of nodes~\cite{hollinger2013sampling}.

The adaptive IPP (AIPP) problem is even more challenging because we seek a reactive policy that chooses the next location to visit based on the observations so far. Initial work on AIPP used information theoretic arguments with myopic heuristics~\cite{stachniss2005information}. A non-myopic approach for AIPP~\cite{singh2009nonmyopic} provides performance guarantees when the utility function is monotone submodular and locations far apart are weakly independent. When the environment task is to narrow down a hypothesis set of possible worlds, an efficient method has been developed with Group Steiner trees~\cite{wee2016adaptive}. Finally, when a prior over informative locations is available, imitation of an oracle achieves good performance while also providing useful theoretical guarantees~\cite{choudhury2017learning,choudhury2018data}.

\subsection{POMDPs}
\label{sec:related-pomdps}
POMDPs provide a principled mathematical framework for sequential decision-making under uncertainty~\cite{kochenderfer2015decision}. A POMDP is typically represented by the tuple $\left(\mathcal{S},\mathcal{A},\mathcal{O},T,Z,R,\gamma\right)$, where $\mathcal{S}$ is the state space, $\mathcal{A}$ is the action space, and $\mathcal{O}$ is the observation space. The transition function $T$ maps states and actions to a distribution over next states, i.e., $T(s,a,s^{\prime}) = P(s^{\prime} \mid s,a)$. When an agent executes an action in a state, it receives a noisy observation of the state modeled by the observation function $Z$, i.e. $Z(s,a,o) = P(o \mid s,a)$.

The reward function $R : \mathcal{S} \times \mathcal{A} \rightarrow \mathbb{R}$ specifies the expected one-step reward received by the agent $R(s,a)$ upon taking action $a$ in state $s$. The discount factor $\gamma \in [0,1)$ is provided for infinite-horizon problems to ensure that the total expected reward over a trajectory $\mathbb{E}[\sum^{\infty}_{t=0} \gamma^t R(s_t,a_t)]$ is bounded when the rewards are bounded. In a POMDP, states are not directly observable. We typically work with the belief space $\mathcal{B}$, which is a space of probability distributions over $\mathcal{S}$. The history of actions and observations can be entirely captured in the current belief state. A POMDP \emph{policy} $\pi : \mathcal{B} \rightarrow \mathcal{A}$ maps the belief state to an action to take.

The objective of solving a POMDP is to obtain an optimal policy, i.e. one that maximizes the expected cumulative reward. Exact solution methods for both finite horizon~\cite{smallwood1973optimal} and discounted infinite horizon~\cite{sondik1978optimal} cases are well-established but intractable in general~\cite{papadimitriou1987complexity}. Consequently, recent efforts have focused on developing approximate solutions~\cite{hauskrecht2000value}.


\subsection{Online POMDP Planning}

Online approaches to POMDPs choose actions at runtime by reasoning over a limited future horizon of belief states reachable from the current belief state. A survey of online methods~\cite{ross2008online} outlines popular approaches, such as branch-and-bound and forward search.

Two recent state-of-the-art methods for online planning are POMCP~\cite{silver2010monte} and DESPOT~\cite{ye2017despot}. The first of these is based on Monte Carlo Tree Search with Upper Confidence Bounds for exploratory actions, while the second uses a sparse approximation of the belief tree rooted at the current belief state. We use the POMCP framework for AIPPMS because it is simpler. Our modifications can also be used with variations of POMCP, such as POMCPOW~\cite{sunberg2018online} for continuous action spaces.
A very recent online approach attempts to address the exploration-exploitation trade-off in informative planning by Pareto-optimal Monte Carlo Tree Search~\cite{chen2019pareto}
but does not allow for multimodal sensing.

\section{Problem Definition}
\label{sec:problem}

The problem that we consider is Adaptive Informative Path Planning with Multimodal Sensing (AIPPMS). We will use notation consistent with previous established work on Adaptive IPP~\cite{singh2009nonmyopic} and its relation to POMDPs~\cite{choudhury2017learning}.

We have a location graph $\graph = \left(\vertexSet, \edgeSet\right)$, where the set of nodes $\vertexSet$ corresponds to all discrete locations the agent can visit and sense. An edge $e = (u,v) \in \edgeSet$ has weight $C(u,v)$ equal to the cost of traveling between locations $u$ and $v$. The robot starts at $\vertexStart$ and needs to reach a given goal node $\vertexGoal$. Each node $\vertex$ has some true \textbf{state} represented by a random variable $\fullWorldSpace_{v}$. The set of all possible states of a node is $\fullWorldSpace$ and the random vector $\world = \left(\worldState{1},\ldots,\worldState{n}\right)$ defines the unobserved true state of the environment. For instance, in the disaster rescue scenario, $\world$ represents the density of survivors at the various pickup locations. The dependencies between location states is modeled by joint distribution $P\left(\world\right)$.

The agent has a suite of sensors $\sensorSet$, with each sensor $\sensor$ having some usage cost $C(\sensor)$.
Let $\meas \in \measSet$ be an \textbf{observation} received by the robot. Let $\measFnDef{}: \vertexSet \times \worldSet \times \sensorSet \to \measSet$ be the observation function. When the robot is at node $\vertex$ in a environment of state $\world$ and uses sensor $\sensor$, the measurement $\meas$ received by the robot is $\meas = \measFnDef\left({\vertex},{\world},\sensor\right)$. The sensor observation model $P\left(\measSet = \meas \mid \vertexSet = \vertex,\sensorSet=\sensor,\world\right)$ is stochastic but the form is known. The sensors typically have varying combinations of performance (fidelity; range) and energy consumption (details in~\Cref{sec:experiments}).

A valid \textbf{action} at the current node $\vertex_t$ is to either go to a different node $\vertex_{t+1}$ or to use some sensor $\sensor$. The movement between locations (and corresponding energy expended) is fully deterministic. The \textbf{cost} of an action $a_t$ taken at current location $\vertex_t$ is therefore either some travel cost to a different node or the energy cost of sensing at that node, i.e.,
\begin{equation}
    C(v_t,a_t) = \begin{cases} C(v_t,v_{t+1}) \ &\text{if} \ a_t = v_{t+1}, \ v_{t+1} \neq v_t \\ C(\sensor) \ &\text{if} \ a_t = \sensor, \ v_{t+1} = v_t \end{cases}.
\label{eq:action-cost}
\end{equation}

Let $\utilityFnDef{}: 2^\vertexSet \times \worldSet \to \real_{\geq 0}$ be a \textbf{utility} function mapping a subset of nodes which have been visited and a world state to some utility which represents the information in the environment. For a collection of visited nodes $\Path$ and a world map encoded in $\world$, $\utilityFn{\Path}{\world}$ assigns a utility. Note that $\utilityFnDef{}$ is a set function. After a node has been visited, visiting it again adds no further utility, but sensing at a visited node may be used as an information gathering action to refine the belief of the true state at other nodes, based on the observation function updates. Given a node $\vertex \in \vertexSet$, a set of nodes $\vertexGroup \subseteq \vertexSet$ and world $\world$, the discrete derivative of the utility function $\utilityFnDef$ is $\marginalGain{\vertex \mid \vertexGroup}{\world} = \utilityFn{\vertexGroup \cup \{ \vertex \}}{\world} - \utilityFn{\vertexGroup}{\world}$.

\begin{figure}[t]
    \centering
    \includegraphics[width=0.8\columnwidth]{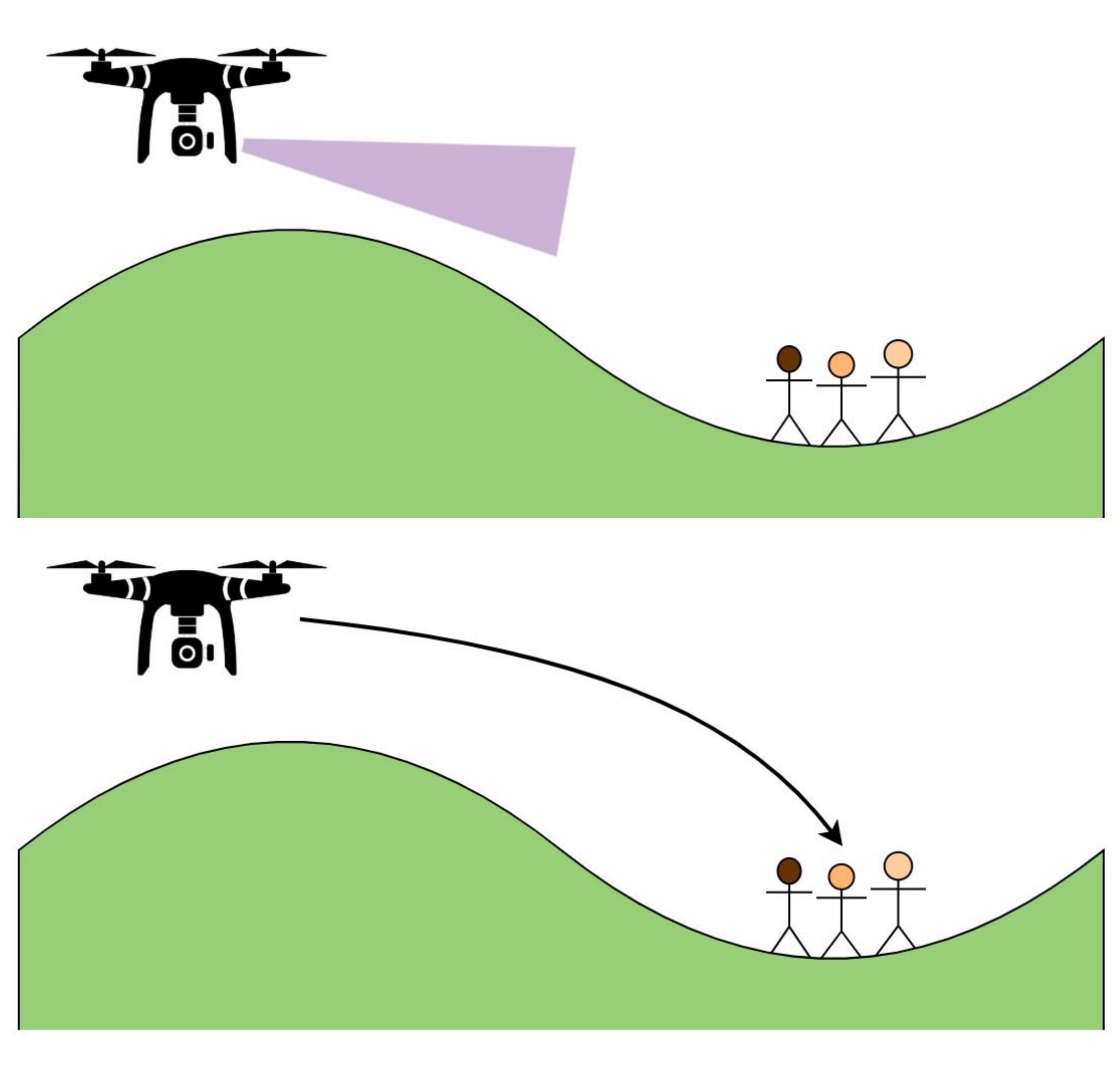}
    \caption{The AIPPMS formulation allows the agent to reason about modulating or completely turning off sensing to conserve energy when it has sufficient information. 
    The upper panel shows how the agent (in a traditional AIPP setting) continues to sense despite sufficient information. The lower panel shows the
    agent directly moving to the next location without potentially expensive sensing.}
    \label{fig:rescue}
\end{figure}

The \textbf{objective} of the general Adaptive IPP problem is to maximize the expected utility of visited nodes, starting at $v_s$ and ending at $v_g$, subject to some cost budget $B$. Note that the \emph{utility} function $\utilityFnDef$ which is relevant to the objective, is entirely distinct from the \emph{cost} function $C$ which is relevant to the constraint. The location selection is done by a policy $\pi$ which, at each timestep $t$, maps the observations received $\{y_i\}_{i=1}^{t-1}$ and locations visited $\{v_i\}_{i=1}^{t-1}$ to the node to visit $v_t$. In our AIPPMS problem, the policy is subject to the same energy constraint, but it can choose to perform a sensing action at timestep $t$ instead of going to another location, while expending some energy cost for sensing. In resource-constrained problems where only a few locations can be visited, sensing can be an important means of determining which locations are likely to have greater utility. \emph{Note that sensing itself has no intrinsic utility in terms of} $\utilityFnDef$.

\section{Approach}
\label{sec:approach}

Our problem inherits all the complexities of Adaptive IPP but adds the additional complexity of deciding between sensors with varying energy costs and observation models. Because of its inherently probabilistic nature, this problem can be framed quite naturally as a POMDP. Despite this applicability, most relevant approaches to AIPP~\cite{singh2009nonmyopic,wee2016adaptive,choudhury2017learning} \emph{do not explicitly solve AIPP as a POMDP}. While each approach states their own reasons for this, an implicit reason is the generality of POMDPs, which does not allow the exploiting of AIPP-specific assumptions. Previous approaches, for example, separately address the utility of an action, which affects the objective, and the feasibility of an action, which affects the energy constraint. 

In the more general AIPPMS, however, decomposability is greatly reduced. The agent must choose between sensing and moving at each timestep, and multiple sensing modalities have different trade-offs between energy and expected information gain. A particularly evocative example is shown in~\Cref{fig:rescue}, where turning off sensing limits energy usage. \emph{We choose to explicitly formulate and solve AIPPMS as a POMDP in order to jointly reason about the effect of movement and multimodal sensing actions}. We describe this formulation in~\Cref{sec:approach-pomdp} and then describe in~\Cref{sec:approach-planning} how we use a online POMDP solver tailored to AIPPMS.

\subsection{POMDP Formulation}
\label{sec:approach-pomdp}

We now formalize AIPPMS as a POMDP using the notation of~\Cref{sec:problem}. The AIPPMS is a discrete-time, finite-horizon, constrained POMDP. The \textbf{state} at time $t$ is defined as $s_t = \left(v_t, \Path_t, \Delta e_t, \world\right)$ where $v_t$ is the agent's current node location, $\Path_t$ is the set of nodes that have been visited already, $\Delta e_t = B - \sum_{i=1}^{t-1}C(s_i,a_i)$ is the remainder of the cost budget. These three components are all fully observable. The world state $\world$ is defined by the specific problem but is not observable. The \textbf{belief} is over the most likely state of the world. In the most general sense, the belief space $\mathcal{B} \subset \worldSet$, but restrictive assumptions on the joint distribution can be made for tractability; \emph{the modeling of the joint distribution between locations is agnostic to the POMDP solver}.

The full \textbf{action} space is the union of all locations and sensors, $\mathcal{A} = \vertexSet \cup \sensorSet$, but the set of valid actions for a state depends on the neighbours of the current node in the location graph, $\mathcal{A}(s_t) = \mathrm{Nbrs}(\vertex_t) \cup \sensorSet$. The \textbf{observation} space is obtained from the AIPPMS specification, $\mathcal{O} = \mathcal{Y}$. An example observation is just the states of some subset of locations, $o_t = \{\hat{\vertexSet},\worldState{\hat{\vertexSet}}\}$ where $\hat{\vertexSet} \subseteq \vertexSet$. For sensing actions, the observations are noisy and depend on the domain-specific sensor observation model. For movement actions, the observation is deterministic and is just the true state of the visited location.

The \textbf{transition} function $T$ is fully deterministic. If $a_t$ is a movement action to a neighboring node, the new state is $s_{t+1} = \left(\vertex_{t+1},\Path_{t+1},\Delta e_{t+1}, \world\right)$ where $\vertex_{t+1} = a_t$, $\Path_{t+1} = \Path_t \cup \vertex_{t+1}$ and $\Delta e_{t+1} = \Delta e_t - C\left(s_t,a_{t+1}\right)$. If $a_t$ is a sensing action, the only difference between $s_t$ and $s_{t+1}$ is for $\Delta e_{t+1} = \Delta e_t - C\left(s_t,a_{t+1}\right)$. The belief state over the world is updated with the observation from the sensor as follows:
\begin{equation}
\label{eq:bel-update}
\begin{split}
    &b_{t+1} = \tau\left(b_t,o_t,\sensor\right) \\
    \implies &b_{t+1}\left(\world^{\prime}\right) \propto \ P\left(o_t \mid \world^{\prime},\sensor, b_t\right) P\left(\world^{\prime} \mid \sensor, b_t\right)\\
    &\propto \ Z\left(\world^{\prime},\sensor,o_t\right) \sum\limits_{\world} T\left(\sensor, \world, \world^{\prime}\right) P(\world \mid b_t) \\
    \implies &b_{t+1}\left(\world^{\prime}\right) \propto \ Z\left(\world^{\prime},\sensor,o_t\right) b_t\left(\world^{\prime}\right)
\end{split}
\end{equation}
which is standard recursive Bayesian estimation of the state of the world. Since the world state is assumed to be fixed, this update can be done more efficiently than in the general case where the state changes. However, the efficiency of this update depends on how the joint distribution over the node states $P(\world)\textbf{}$ is maintained; in general, MAP inference with belief networks is NP-hard~\cite{shimony1994finding}.

The \textbf{reward} function for the POMDP is defined in terms of the AIPPMS utility function $\utilityFnDef$ as follows:
\begin{equation}
\label{eq:reward}
R(s_t, a_t, s_{t+1}) = \begin{cases} 0 & \text{if } a_t = \sensor \\ \marginalGain{\vertex_{t+1} \mid \Path_t}{\world} & \text{if } a_t = \vertex_{t+1} \end{cases}
\end{equation}
Therefore, an agent receives reward when it visits a node and observes its true underlying state.
This reward is different from the action's cost, which is obtained from~\eqref{eq:action-cost}.
The problem terminates when the agent has too little energy left to take another action, i.e. $\Delta e_t < \min_{a} C(s_t,a)$. If the agent is not at the goal $\vertexGoal$ when this happens, it receives a reward of $-\infty$. If it is at the goal, the cumulative reward is the utility of all the visited nodes.
The deterministic nature of the energy cost function allows us to define a state-dependent feasible set of actions that do not violate the energy constraint.
We will discuss subsequently how we incorporate this feasible action set in our specific online planning framework.

\subsection{Online Planning for AIPPMS}
\label{sec:approach-planning}

We have described in~\Cref{sec:related-pomdps} the computational challenges involved in solving POMDPs, and the motivation for online planning in large domains with substructure in the reachability of states. For AIPPMS, the connectivity of the graph $\graph$ makes only certain states reachable from other states (the possible values of $\vertex_{t+1}, \  \Path_{t+1}, \text{ and } \Delta e_{t+1}$ given their values at $t$ are restricted by the graph). Furthermore, both state and observation spaces are exponential in the number of nodes, making an online planning approach preferable.

We use Partially Observable Monte Carlo Planning or POMCP~\cite{silver2010monte} as the underlying online solver. We tailor POMCP to solve AIPPMS problems with two specifications, that we now
describe. First, we prune all constraint-violating actions during the lookahead search from the current state. Second, we develop a rollout policy that is quite suitable for a relevant class of utility functions.

\subsubsection{Action Pruning}

The constraint for AIPPMS is to reach the goal vertex $\vertex_g$ within the cost budget $B$. Due to the deterministic behavior of the constraint cost, we can exactly specify a feasibility condition on any state of the POMDP. For the location graph $\graph$, let the shortest path between any two vertices $u$ and $v$ on the graph be denoted as $C_{\graph}(u,v)$. Then, a state $s_t = \left(\vertex_t,\Path_t,\Delta e_t,\world\right)$ is feasible if $\Delta e_t > C_{\graph}(u,v)$, i.e. the agent has sufficient energy in that state to go to the goal.

\begin{figure}[t]
    \centering
    \begin{subfigure}{0.8\columnwidth}
        \centering
        \includegraphics[width=\textwidth]{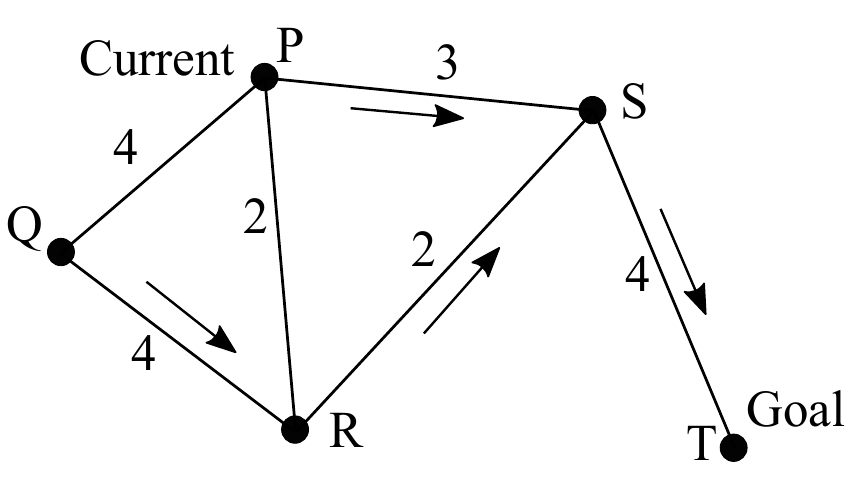}
        \caption{}
        \label{fig:constr-scenario}
    \end{subfigure}
    \begin{subfigure}{0.8\columnwidth}
        \centering
        \includegraphics[width=\textwidth]{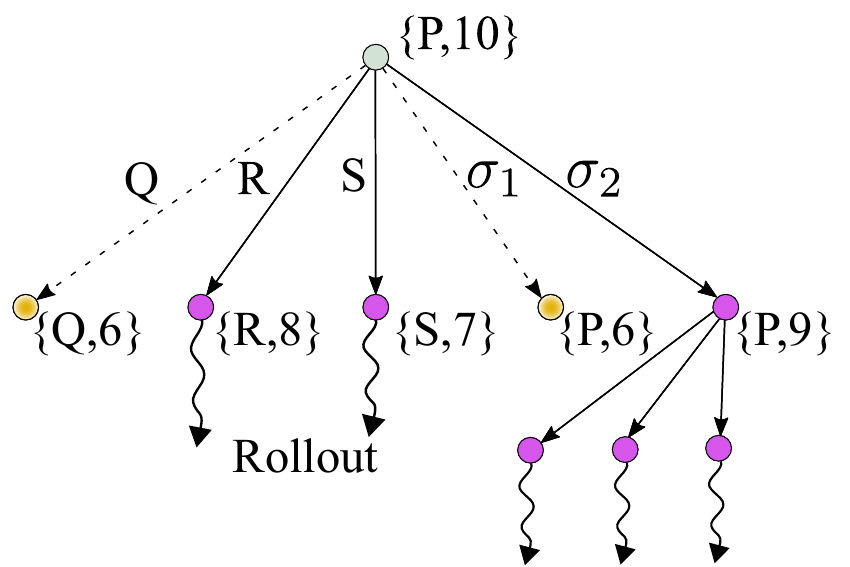}
        \caption{}
        \label{fig:constr-search}
    \end{subfigure}
    \caption{Constrained action selection in our modified POMCP. (\subref{fig:constr-scenario})\textbf{} An example scenario with arrows representing the direction of the shortest cost path. (\subref{fig:constr-search}) Action pruning resulting from the energy budget -- dotted line actions are never taken, while rollouts continue down bold paths.}
    \label{fig:constr}
\end{figure}

Denote the set of all feasible states as $\bar{\mathcal{S}}$. Then, for any feasible state $s \in \bar{\mathcal{S}}$, the set of feasible actions comprises those that can only lead to another feasible state, i.e.
\begin{equation}
\label{eq:feas-act}
\bar{\mathcal{A}}(s) = \{a \in \mathcal{A} \mid T(s,a,s^{\prime}) > 0 \Rightarrow s^{\prime} \in \bar{\mathcal{S}} \}
\end{equation}
and can be computed efficiently for any state by caching and looking up the all-pairs shortest paths matrix for $\mathcal{G}$ using Floyd-Warshall's algorithm~\cite{floyd1962algorithm}. An illustration with a particular scenario is shown in~\Cref{fig:constr}. The action pruning enables more exhaustive searches for the same planning time.

\subsubsection{GCB Rollout Policy}
For POMCP, the rollout policy is used in the second stage of simulations to estimate the value of a leaf node in the search tree. In principle, the rollout policy could be an uninformed one that chooses actions at random, but in practice, the choice of rollout can greatly impact the performance of POMCP on large problems.

\begin{algorithm}[t]
\caption{POMCP with GCB Rollout for AIPP-MS}
\begin{algorithmic}[1]
\Procedure{Simulate}{$s$, $u$, $depth$}
\If {$\gamma^{depth} < \epsilon$}
    \State \textbf{return} 0
\EndIf
\If {$h \notin T$}
    \ForAll {$a \in \bar{\mathcal{A}}(s)$}
        \State $T(ha) \leftarrow \left(N_{init}(ha),V_{init}(ha),\varnothing \right)$
    \EndFor
    \State \textbf{return} \textsc{Rollout}$(s,h,depth)$ \Comment{This uses \textsc{Action} internally}
\EndIf
\State $a \leftarrow \underset{{ b \ \in \bar{\mathcal{A}}(s)} }{\mathrm{argmax}} \ V(hb) + c\sqrt{\frac{\mathrm{log} \ N(h)}{N(hb)}}$
\State $\left(s^{\prime},o\textbf{},r\right) \sim \mathcal{G}_{\text{sim}}(s,a)$
\State $R \leftarrow r + \gamma \cdot \textsc{Simulate}\left(s^{\prime},hao,depth+1\right) $
\State $B(h) \leftarrow B(h) \cup \left\{s\right\}$
\State $N(h) \leftarrow N(h) + 1$
\State $N(ha) \leftarrow N(ha) + 1$
\State $V(ha) \leftarrow V(ha) + \frac{R - V(ha)}{N(ha)}$
\State \textbf{return} $R$
\EndProcedure
\Statex
\Procedure{Action}{$\pi_{rollout}$,$s$,$h$} \Comment{GCB Rollout}
\ForAll{$a \ \in \bar{\mathcal{A}}(s)$}
    \If{$a \ \in \vertexSet$}  \Comment{$a$ is for movement}
        \State $U(a) \leftarrow \mathbb{E}_{b(\world)}\left[\marginalGain{a \mid \Path}{\world}\right] / C(s,a)$
    \Else \Comment{$a$ is for sensing}
        \State $U(a) \leftarrow \mathrm{IG}\left(a \mid b(\world)\right) / C(a)$
    \EndIf
\EndFor
\State \textbf{return} $a \sim \mathrm{SoftMax}(U)$
\EndProcedure
\end{algorithmic}
\label{alg:pomcp-mod}
\end{algorithm}

We use a rollout policy based on the Generalized Cost-Benefit or GCB Algorithm~\cite{zhang2016submodular}, which is designed for the fully observable IPP problem (when all utilities of visiting locations are known). The performance guarantees of GCB only hold for problems where the utility function is submodular, i.e. obeys the property of diminishing returns~\cite{nemhauser1978analysis}. This is not an issue for us because submodular functions are quite prevalent in real-world settings~\cite{krause2014submodular} and particularly in informative path planning~\cite{meliou2007nonmyopic,binney2010informative}. The key idea of GCB is to greedily choose the next action or option that maximizes the ratio of marginal utility or benefit to cost expended. For our partially observable stochastic optimization case, we use an adaptive greedy strategy that computes expected marginal utility~\cite{golovin2011adaptive}. 

The adaptive greedy rollout policy is outlined in the ACTION procedure in~\Cref{alg:pomcp-mod}. As before, we only consider the set of feasible actions for the state. For movement actions ($a \in \vertexSet$), the expected marginal utility is computed based on the belief state $b(\world)$ which is equivalently encoded in the history $h$. Sensing actions ($a \in \sensorSet$) have no utility with respect to the AIPPMS objective $\utilityFnDef$, but we incentivize them in the rollout with an information-theoretic reward,
\begin{equation}
\label{eq:ig}
\begin{split}
\text{IG}(a|b) 
&= \sum_o P(o \mid b, a) \left[ \max_s \tau(b,o,a)(s) - \max_s b(s) \right] \\
&\approx \sum_i^{\text{\# samples}} \left[ \max_s \tau(b,o_i,a)(s) - \max_s b(s) \right]
\end{split}
\end{equation}
Sensing actions are thus selected to maximize in expectation the mode of the belief state. This technique combines an expected information gain approach~\cite{stachniss2005information} with the insight that distribution mode can be used as a lightweight approximation of negative information entropy~\cite{dressel2017efficient}, as collapsed distributions necessarily have more concentration of density. 
Finally, having computed the expected cost-benefit ratio for each of the feasible actions from the state, we sample actions from a softmax distribution over these values.

\section{Experiments}
\label{sec:experiments}

We run all simulations in the Julia programming language for its fast numerical computations~\cite{bezanson2017julia}. We used the POMDPs.jl framework for specifying our POMDP formulation and for the base implementation of POMCP~\cite{egorov2017pomdps}.

\subsection{Baselines}
\label{sec:experiments-baseline}

Since we are introducing AIPPMS in this work, there is no existing baseline we could compare against directly. We therefore extend the NAIVE or Nonmyopic Adaptive InformatiVE path planning algorithm~\cite{singh2009nonmyopic}, which is an elegant and theoretically motivated approach for AIPP, to our AIPPMS problem. The NAIVE algorithm consists broadly of iterative Bayesian updating (the same as for the modified POMCP) and replanning using a nonadaptive method called \emph{pSPIEL-Orienteering} or $\mathrm{PSPIEL}_{OR}$. We omit an elaboration of NAIVE here and refer readers to its paper. \emph{NAIVE is the acronym of the algorithm and is not meant to indicate that the baseline is actually naive}.

In Algorithm 1 of the reference work for NAIVE, at each iteration, the non-adaptive $\mathrm{PSPIEL}_{OR}$ method plans a path $\mathcal{P}_t$ on the graph $\graph$ from the current location to the goal within the remaining budget. There is no sensing because the AIPP problem has no notion of multimodal sensors. Since AIPPMS has the same state space as AIPP, we can compute $\mathcal{P}_t$ using $\mathrm{PSPIEL}_{OR}$ at each iteration in modified NAIVE as well. We compute the expected utility of $\mathcal{P}_t$ as follows
\begin{equation}
    U(\mathcal{P}_t) = \sum\limits_{v \in \mathcal{P}_t} \mathbb{E}_{b_t(\world)} [\marginalGain{v \mid \Path_t}{\world}]
\end{equation}
which is an efficient approximation of the true expected utility of the path given the current belief state $b_t(\world)$ and set of visited nodes $\Path_t$. In addition, we compute the best expected information gain among the feasible sensing actions,
\begin{equation}
    U^{*}_{\sensorSet}(b_t) = \underset{\sensor \in \bar{\mathcal{A}}(s_t)}{\mathrm{max}} \ \mathrm{IG} \left(\sensor \mid b_t(\world)\right) 
\end{equation}
where $\sensor_{t}^{*}$ is the corresponding sensor.
Finally, we choose the next action to take by comparing two scaled utilities,
\begin{equation}
    a_t = \begin{cases} \mathcal{P}_t[2] \ &\text{if } \lambda \cdot U(\mathcal{P}_t) > (1-\lambda) \cdot U^{*}_{\sensorSet}(b_t) \\ \sensor_{t}^{*} \ &\text{otherwise } \end{cases}
\end{equation}
where $\lambda \in [0,1]$ is a tuning parameter that prioritizes exploration through sensing when close to 1 and exploitation by moving when close to 0. For an ablation study of the benefit of the GCB rollout, we will also compare against POMCP using a random policy to choose
actions during rollout.

\begin{figure}
    \centering
    \includegraphics[width=\columnwidth]{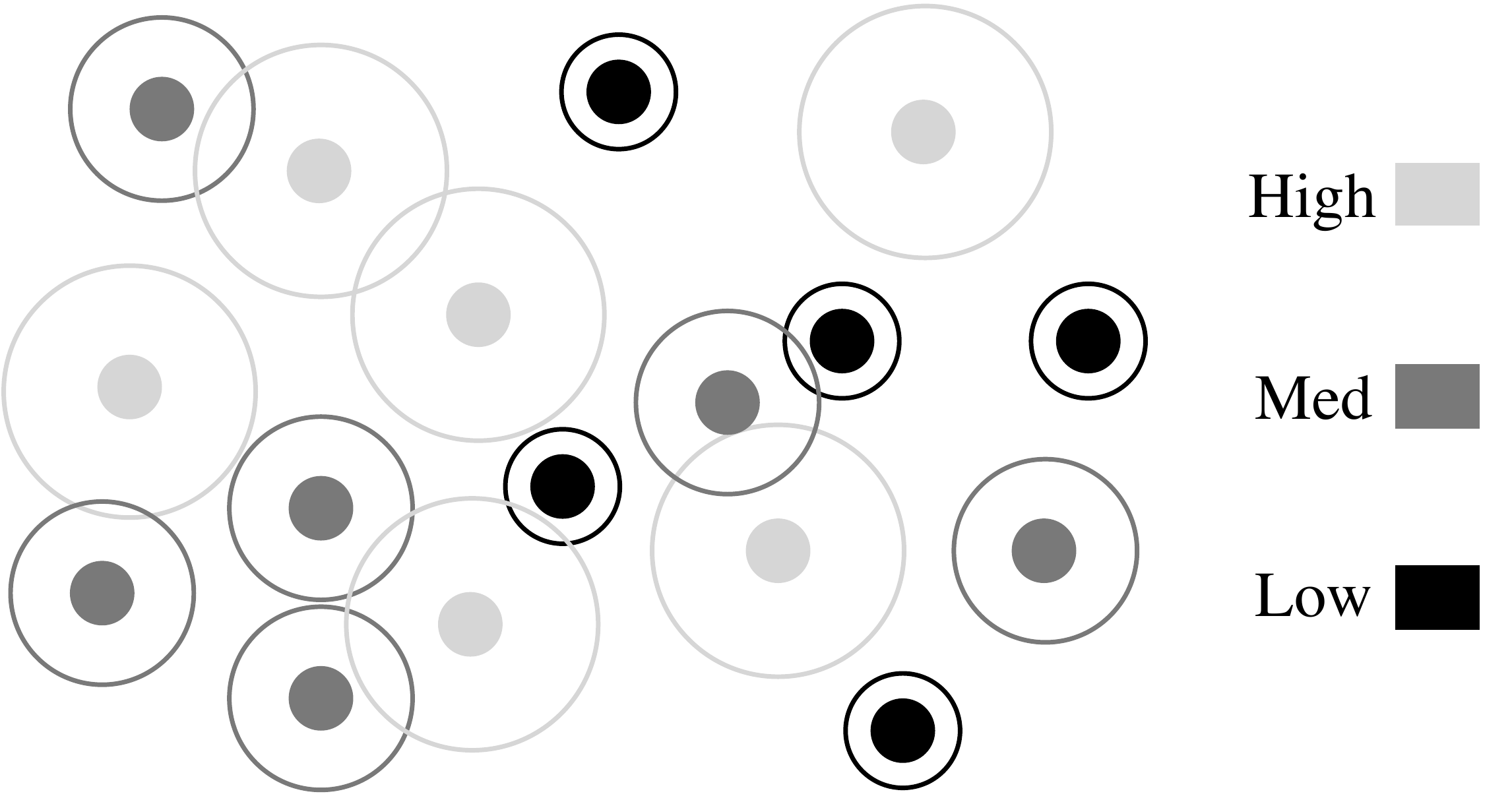}
    \caption{An illustration of the utility in the search-and-rescue problem scenario used for our experiments. Each node in the graph has a state corresponding to one of high, medium or low accessibility to survivors. The utility of visiting a node is the additional area covered by the circle, given what has been visited already. The utility function in this setting is therefore a monotone submodular set function.}
    \label{fig:gridcov}
\end{figure}

\subsection{Domain 1: Search-and-Rescue}
\label{sec:experiments-search}

Our first experimental domain is based on the search-and-rescue scenario that was presented with the original NAIVE algorithm. In the aftermath of a disaster, survivors are scattered uniformly across some terrain. The agent is an aerial vehicle with limited energy that can visit a number of locations and rescue however many people are stranded there (depending on the accessibility of a particular location to the survivors).

The environment is represented as a unit grid, on which a location graph is generated by sampling points as nodes and joining two nodes with an edge if their mutual distance is less than some threshold $\rho$. The cost of traversing an edge is the scaled Euclidean distance between the two endpoint nodes. Each node or location is \emph{independently} assigned a true state that dictates the utility of visiting it. The independence assumption is not one made by our POMCP-based approach, but allows a fairer comparison with modified NAIVE, which assumes locality (implied by independence).

For our search-and-rescue setting, there are three types of true states corresponding to low, medium, and high accessibility of the locations to survivors, represented by the area covered by the location (see~\Cref{fig:gridcov}). The marginal utility of a location in a particular state, given the locations visited already, is calculated by discretizing the grid and computing the union of tiles covered by the location in that state. The key difference in our environment compared to that for NAIVE is the existence of multiple sensor models, which offer noisy observations of the true states at various locations. The sensors are modeled by a maximum fidelity parameter $A$, fidelity decay rate $r$, and energy cost $C$. Each usage of sensor $\sensor$ incurs cost $C(\sensor)$ and yields a correct observation of the underlying state of a node with probability $P(o_i = s_i \mid s_i, \sensor) = A_{\sensor}\cdot r_{\sensor}^d$
where $d$ is the distance between the agent and the node. Other incorrect observations have uniform likelihood given an incorrect sensor reading. Thus, $P(o_i \neq s_i \mid s_i) = (1 - A_{\sensor}\cdot r_{\sensor}^d)/2$.

\subsubsection{Results}
\label{sec:experiments-search-results}

We compared our modified POMCP (with GCB Rollout) to POMCP with Random Rollout and modified NAIVE on problems randomly generated according to the search-and-rescue scenario described above. For each problem, the location graph $\graph$ had $30$ nodes and the radius threshold $\rho$ for edges was sampled from $\mathrm{U}(0.25,0.4)$. We ensured via rejection sampling that each generated graph had exactly one connected component. We randomly sampled a node from $\graph$ and set it to be both the start $\vertex_s$ and goal $\vertex_g$. Therefore, the agent has to return to its starting point after gathering information. For each problem, we set the budget to approximately two-thirds of the Traveling Salesman cost on the graph from the start node, thereby ensuring that not all nodes could be visited and incentivizing at least some sensing.

\begin{table}
\caption{Both POMCP variants significantly outperform NAIVE on three different utility distributions in the environment. The GCB Rollout
is consistently better than random, though the relative gap is not very large. Rewards were averaged over $30$ different trials
for each setting.}
\centering
\begin{tabular}{@{} lrrr @{}}
  \toprule
  Hi, Med, Lo & NAIVE & POMCP & POMCP\\
  Distribution & & Random & GCB\\
  \midrule
  $[1/6, 1/6, 2/3]$  & $295.7$  & $480.0$   & $\mathbf{555.3}$  \\
  $[1/3, 1/3, 1/3]$  & $628.3$  & $993.0$   & $\mathbf{1020.0}$  \\
  $[2/3, 1/6, 1/6]$  & $1104.7$ & $1341.3$  & $\mathbf{1509.0}$  \\
  \bottomrule
\end{tabular}
\label{table-area2d}
\end{table}

We considered three different sets of problems, based on the distribution of accessibility types (high, medium, low) that node states are independently sampled from. 
For each setting, we generated $30$ different problems and ran all approaches on them. ~\Cref{table-area2d} depicts the 
average utility or reward obtained by the approaches. All algorithms are guaranteed to be feasible by construction, so we only focus on the utility gathered.
\emph{The magnitude of the average utility is a function of the grid discretization; the relative performance is truly of interest}. 

Over the problem sets, our modified POMCP with GCB Rollout consistently accrues the most utility, significantly outperforming modified NAIVE and also POMCP with Random Rollout, albeit to a lesser extent (the relative performance gap between GCB and Random Rollout increases
on a more challenging problem in~\Cref{sec:experiments-isrs}).
As expected, with an increase in the proportion of the higher utility states, the absolute utility accrued by all approaches increases.
More notably, the \emph{relative performance gap is highest for the first set}, where high utility states are least prevalent. This suggests that the modified POMCP balances sensing and movement more effectively than modified NAIVE when identifying which states are of likely higher or lower utility.
The average iteration of NAIVE requires \SI{1}{\second} of computation and that of POMCP requires \SI{6}{\second}. Both are reasonable for our purposes, and increasing the computation time for NAIVE through the relevant parameters did not improve performance.

\emph{Fundamental differences between a POMCP approach and NAIVE explain the performance gap on AIPPMS, over and above implementation quality}. Each candidate path computed by modified NAIVE ignores sensing actions. Subsequently, NAIVE computes the expected information gain of possible sensing actions from the current belief state and then compares that with the candidate path to decide whether to move or sense. However, the lookahead search in POMCP can simulate the effects of movement followed by sensing, and sensing followed by movement. Therefore, it can identify some good future sequences of potentially \emph{interleaved sensing and movement actions}, and then decide which is the next best action to take.

\begin{figure}[t]
    \centering
    \includegraphics[width=0.6\columnwidth]{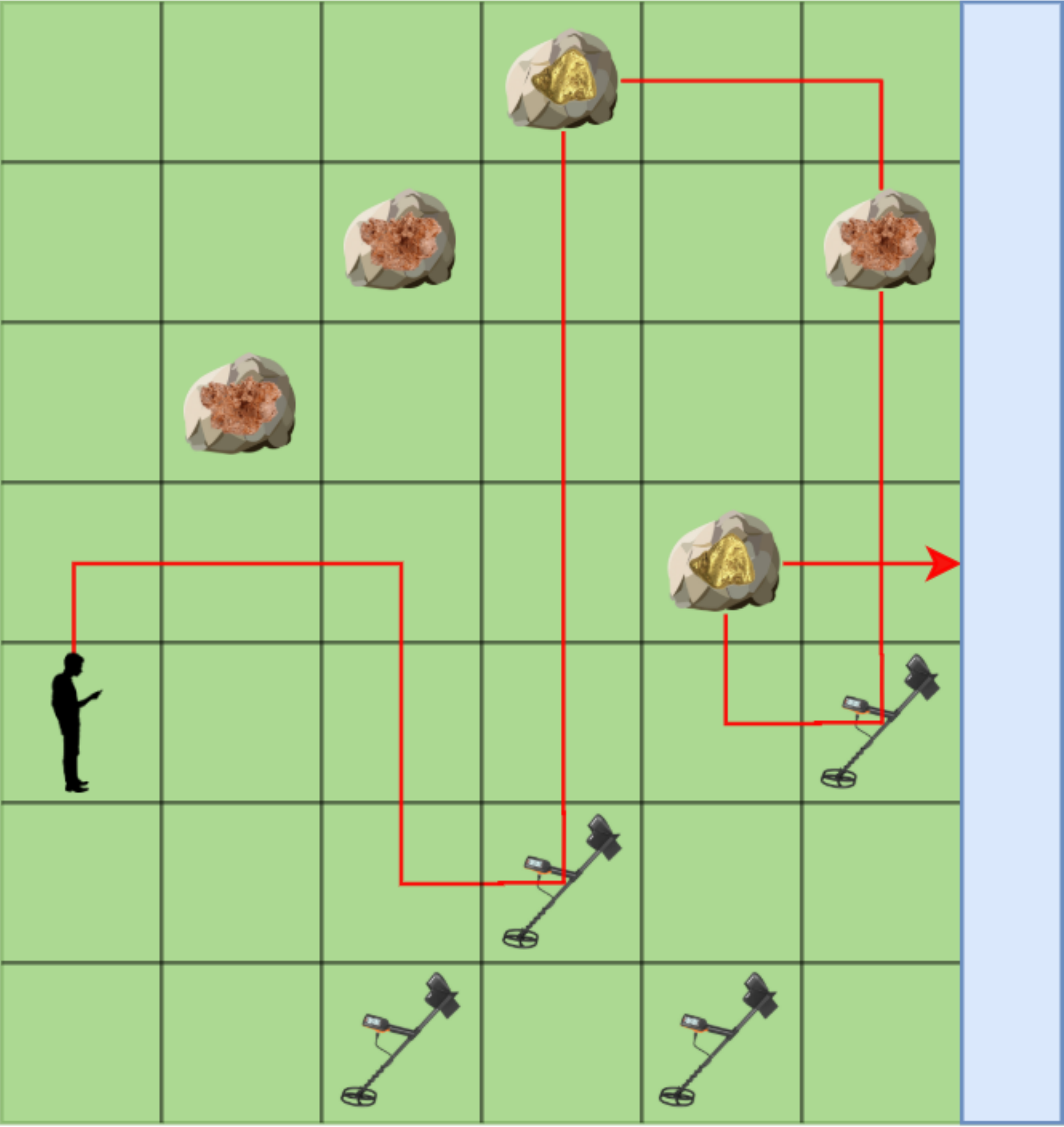}
    \caption{An illustration of the Information Search RockSample problem, our second evaluation domain. The agent uses beacons to sense the states of rocks, and gathers utility
    by visiting good rocks.}
    \label{fig:isrs-viz}
\end{figure}

\begin{table*}[th]
\caption{On the Information Search Rock Sample domain, the GCB rollout significantly outperforms the random 
rollout over all settings by making better use of beacons and visiting more good rocks.
We averaged the rewards from $50$ different trials for each setting. The standard error of the mean
is less than $10\%$ of the mean in each case.}
\centering
\begin{tabular}{@{} llrrcrrcrr @{}}
  \toprule
  Rocks & Beacons & \multicolumn{2}{c}{$p = 0.5$} & \phantom{a} & \multicolumn{2}{c}{$p = 0.75$} & \phantom{a} & \multicolumn{2}{c}{$p = 1.0$}\\
  \cmidrule{3-4} \cmidrule{6-7} \cmidrule{9-10} $k$ & $b$   & Random  & GCB       & & Random   & GCB && Random & GCB\\
  \midrule
  $10$  & $10$  & $21.6$    & $\mathbf{29.4}$    & & $25.2$  & $\mathbf{38.2}$    & & $30.0$ & $\mathbf{49.0}$ \\
  $10$  & $25$  & $23.4$    & $\mathbf{27.8}$    & & $26.8$  & $\mathbf{41.0}$    & & $24.2$ & $\mathbf{47.4}$ \\
  $25$  & $10$  & $45.0$    & $\mathbf{63.6}$    & & $54.2$  & $\mathbf{87.8}$    & & $63.6$ & $\mathbf{121.8}$ \\
  $25$  & $25$  & $41.8$    & $\mathbf{77.0}$    & & $52.6$  & $\mathbf{105.0}$   & & $69.4$ & $\mathbf{120.8}$ \\
  \bottomrule
\end{tabular}
\label{table-isrs}
\end{table*}

\subsection{Domain 2: Information Search RockSample}
\label{sec:experiments-isrs}

To further motivate the AIPPMS formulation and evaluate our approach, we also adapt the Information Search RockSample domain or ISRS~\cite{he2011efficient}.
It is a variant of the classical RockSample problem for POMDPs that is both far more challenging
and more suitable for comparing informative path planning algorithms. Briefly, the ISRS problem models a rover exploring an unknown
terrain, represented as an $n \times n$ grid. Scattered over the grid are $k$ rocks, with at most one rock per grid cell. Only some
of the rocks are `good', i.e. have scientific value and yield a positive reward. Once a rock is visited it becomes `bad'
and provides no further reward when sampled. The positions of the rover and rocks
are known apriori, but only visiting a rock reveals its state. See~\Cref{fig:isrs-viz} for an illustration.

There is also a set of $b$ beacons (one per cell, no overlap with rock locations) in the grid. The rover \emph{must visit the beacons
in order to take sensing actions} and get observations about the state of the nearby rocks, where the fidelity of the observation
reduces with increasing distance between the beacon and the corresponding rock. As before, there are multiple sensor modalities
with complementary trade-offs of usage energy and fidelity parameters. Moving between adjacent cells also expends energy cost. The rover
must return to the origin cell without exceeding its energy budget.

\subsubsection{Results}
\label{sec:experiments-isrs-results}

For ISRS, we focus on the interplay between three problem parameters: the number of rocks $k$, the number of beacons $b$, and the relative proportion of good rocks, through the independent Bernoulli probability $p$ of a rock
being good. Accordingly, we vary these three parameters for our experiments while keeping the others fixed. We set the size of the grid to $10 \times 10$, the energy budget to be $100$ units, where the energy cost of each movement is $1$ unit and that of using the two sensors are $0.5$ and $2$ units respectively, and the reward for sampling a good rock to be $10$ units.

~\Cref{table-isrs} compares the average reward for POMCP with the random rollout strategy to that for POMCP with our
GCB rollout strategy, over a range of $\{k,b,p\}$ settings, averaged over $30$ trials for
each setting. NAIVE accrued little to no reward for most settings and trials,
so we omitted it in the interest of space. This behavior of modified NAIVE is not too surprising. For the ISRS domain with its beacons, explicitly reasoning about the trade-off between movement and multimodal sensing
is particularly important for good performance. \emph{Therefore, the ISRS domain provides strong empirical justification for extending
the AIPP formulation to incorporate multimodal sensing, in addition to our earlier justification from first principles.}

We highlight \textbf{three key observations} from the readings in~\Cref{table-isrs}. First and foremost, GCB consistently
outperforms Random across all settings, far more so than it did for the search-and-rescue domain.
This finding further underscores how ISRS is significantly more challenging than search-and-rescue.
Second, the relative performance gap of GCB to Random increases both with more good rocks (higher $k$) and with a higher proportion
of good rocks (higher $p$), e.g. compare the relative performance for $\{10,10,0.5\}$ to both $\{10,10,0.75\}$ and to $\{25,25,0.5\}$. Third, for the same $k$ (rocks) but with increasing $b$ (beacons), the performance of GCB relative
to itself either stays the same or increases, e.g., compare GCB for $\{25,10,0.75\}$ to $\{25,25,0.75\}$, while for Random it does not increase appreciably for any setting. This supports the intuitive hypothesis that GCB is making better use
of environmental information to improve its estimate of the goodness of a rock.

\section{Conclusion}
\label{sec:conclusion}

We extended the Adaptive Informative Path Planning problem to a setting with Multimodal Sensing (AIPPMS). In contrast to previous AIPP approaches that eschew a POMDP formulation as being too general and intractable, we embraced POMDPs as the appropriate structure to jointly reason about movement and sensing for the more general AIPPMS problem. Due to the large state and observation space and implicit reachability structure of AIPPMS, we used the online planning framework of Partially Observable Monte Carlo Planning, with modifications to the action selection and an adaptive greedy rollout policy based on Generalized Cost-Benefit. Our resulting approach consistently outperforms the modified NAIVE algorithm over multiple domains, and our rollout policy is a key contributor to this performance.

Future research could address many of our limitations. We take an empirical approach in this paper
in contrast to most work on AIPP that conducts theoretical analyses (albeit under modeling assumptions). A more rigorous approach to analyzing AIPPMS, under appropriate assumptions on the utility and sensor models, would be of interest and may precipitate the development of high-performance tailored algorithms. As we motivated earlier, we used the POMCP framework for its simplicity and ease of extension, but an extensive study of other state-of-the-art solvers, both offline and online, would be instructive.
In our AIPPMS formulation, the inter-dependency between movement and sensing actions is weak; the costs and utilities of the actions of each type are unaffected by actions of the other type. Perhaps the POMDP approach would have even more substantial gains in performance if the coupling between action types was stronger (this would require different utility and cost models). Finally, given the greater applicability of AIPPMS over AIPP to real-world robotics problems with energy costs for sensing, implementing the modified POMCP for such an application would be appropriate.

\bibliographystyle{aaai}
\bibliography{refs}

\end{document}